\documentclass[runningheads]{llncs}
\usepackage{graphicx}
\usepackage{amsmath}
\usepackage{bm}
\usepackage{gensymb}
\usepackage{amssymb}
\usepackage{siunitx}
\usepackage{microtype}
\usepackage{xspace}
\usepackage{xcolor}

\usepackage{booktabs}
\usepackage{varioref}
\usepackage{hyperref}
\hypersetup{
pagebackref=true,breaklinks=true,colorlinks,bookmarks=false,citecolor=green!80!black,linkcolor=red!80!black,urlcolor=blue
}
\usepackage{cleveref}
\usepackage{subcaption}
\usepackage{tikz}
\usepackage{pgfplots}

\crefname{section}{Sec.}{Sections}
\crefname{figure}{Fig.}{Figure}
\crefname{table}{Tab.}{Table}
\crefname{equation}{Equ.}{Equation}

\usepackage[inline]{enumitem}
\setlist*[enumerate]{label=(\arabic*)}

\newcommand{\onedot}{.\xspace}
\newcommand{\etal}[1]{#1~et~al\onedot}

\newcommand{\eg}{e.\,g.,\xspace}

\newcommand{\cf}{cf\onedot}
\newcommand{\ie}{i.\,e.,\xspace}
\newcommand{\wrt}{w.\,r.\,t\onedot}

\newcommand{\img}{\bm{I}}
\newcommand{\iimgx}{\img_{\Sigma x}}
\newcommand{\iimgy}{\img_{\Sigma y}}
\newcommand{\diimgy}{\nabla_\sigma\img_{\Sigma y}}
\newcommand{\diimgx}{\nabla_\sigma\img_{\Sigma x}}
\newcommand{\diimgk}{\nabla_\sigma\img_{\Sigma k}}
\newcommand{\bc}{\bm{b}}
\newcommand{\mc}{\bm{m}}
\newcommand{\xc}{\bm{x}}
\newcommand{\xh}{\tilde{x}}

\newcommand{\mch}{\tilde{\bm{m}}}
\newcommand{\xch}{\tilde{\bm{x}}}
\newcommand{\homo}{\bm{H}}

\newcommand{\tp}{\intercal}

\newcommand{\dataa}{\textsc{DataA}\xspace}
\newcommand{\datab}{\textsc{DataB}\xspace}
\newcommand{\datac}{\textsc{DataC}\xspace}
\newcommand{\datad}{\textsc{DataD}\xspace}

\begin{document}
%
% TODO: wuerde das alles gross schreiben:
% Fast and Robust Detection of Solar Modules for Electroluminescence Imaging
\title{Fast and robust detection of solar modules in electroluminescence images}
\author{Mathis Hoffmann\inst{1,2} \and
Bernd Doll\inst{2,3,4} \and
Florian Talkenberg\inst{5} \and
Christoph J. Brabec\inst{2,3} \and
Andreas K. Maier\inst{1} \and
Vincent Christlein\inst{1}}
\authorrunning{M.\ Hoffmann et al.}
\institute{Pattern Recognition Lab, Friedrich-Alexander-Universit\"at Erlangen-N\"urnberg, Germany \and
Institute Materials for Electronics and Energy Technology, Friedrich-Alexander-Universit\"at Erlangen-N\"urnberg, Germany
\and
Helmholtz-Institut Erlangen-N\"urnberg, Germany
\and
Graduate School in Advanced Optical Technologies, Erlangen, Germany
\and
greateyes GmbH, Berlin, Germany
}
\maketitle              % typeset the header of the contribution
\begin{abstract}

Fast, non-destructive and on-site quality control tools, mainly high sensitive imaging techniques, are important to assess the reliability of photovoltaic plants. To minimize the risk of further damages and electrical yield losses, electroluminescence (EL) imaging is used to detect local defects in an early stage, which might cause future electric losses.
For an automated defect recognition on EL measurements, a robust detection and rectification of modules, as well as an optional segmentation into cells is required. This paper introduces a method to detect solar modules and crossing points between solar cells in EL images. We only require 1-D image statistics for the detection, resulting in an approach that is computationally efficient. In addition, the method is able to detect the modules under perspective distortion and in scenarios, where multiple modules are visible in the image. We compare our method to the state of the art and show that it is superior in presence of perspective distortion while the performance on images, where the module is roughly coplanar to the detector, is similar to the reference method. Finally, we show that we greatly improve in terms of computational time in comparison to the reference method.

\end{abstract}
\section{Introduction}

\begin{figure}[t]
    \centering
    \includegraphics[width=.25\linewidth,trim=170 0 150 0,clip,angle=90]{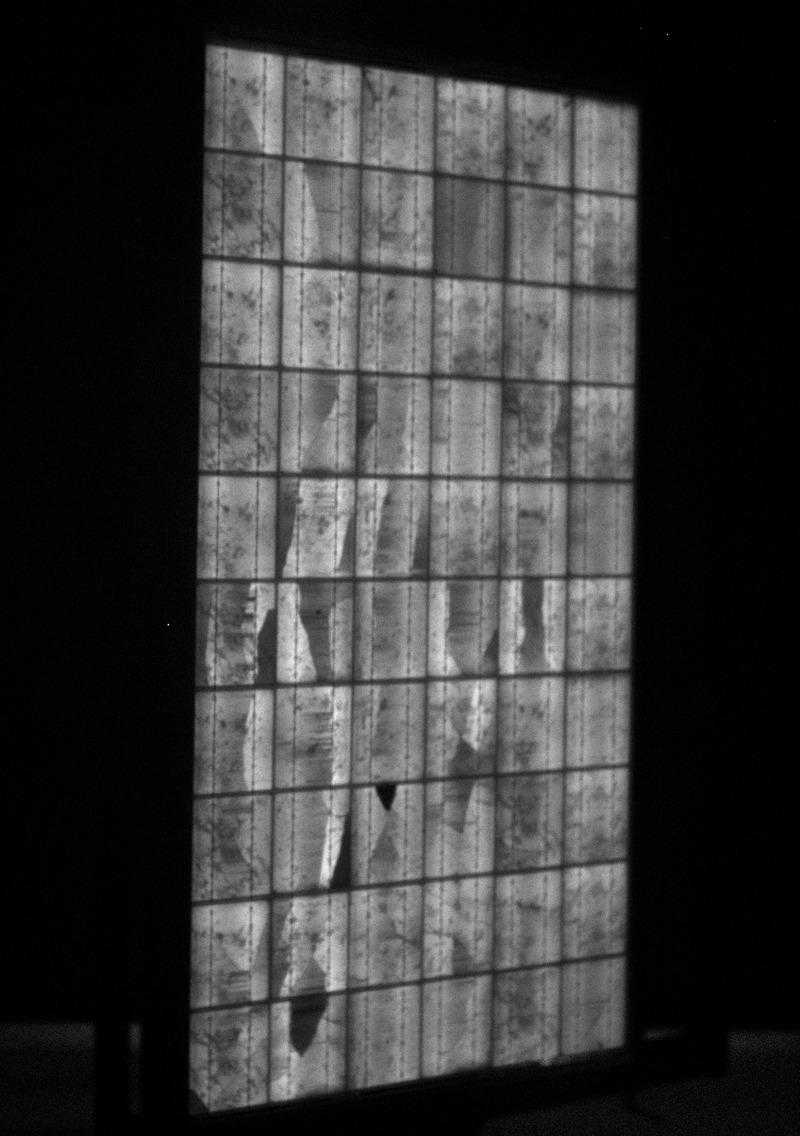}
    \caption{Example EL image.}
    \label{fig:example-module}
\end{figure}

Over the last decade, photovoltaic (PV) energy has become an important factor in emission-free energy production. 
In 2016 for example, about \SI{40}{\giga\watt} of PV capacity was installed in Germany, which amounts to nearly one fifth of the total installed electric capacity~\cite{EC2018}. Not only in Germany, renwable elecricity production has been transformed to a considerable business. It is expected, that by 2023 about one third of world wide electricity comes from renwable sources~\cite{IEA2018}.
To ensure high performance of the installed modules, regular inspection by imaging and non-imaging methods is required. For on-site inspection, imaging methods are very useful to find out which modules are defect after signs of decreasing electricity generation have been detected. Typically, on-site inspection of solar modules is performed by infrared (IR) or electroluminescence (EL) imaging. This work focusses on EL imaging. However, it could be adapted to other modalities as well.

A solar module (see \cref{fig:example-module}) consists of a varying number of solar cells that are placed onto a regular grid. Since cells on a module share a similar structure and cracks are usually spread out only within each cell, it is a natural algorithmic choice to perform detailed inspection on a per cell basis. To this end, an automatic detection of the module and crossing points between cells is required.

Our main contributions are as follows: We propose a method for the detection of solar modules and the crossing points between solar cells in the image. It works irrespective of the module's pose and position. Our method is based on 1-D image statistics, leading to a very fast approach. In addition, we show how this can be extended to situations, where multiple modules are visible in the image. Finally, we compare our method to the state of the art and show that the detection performance is comparable, while the computational time is lowered by a factor of \num{40}.

The remainder of this work is organized a follows: In \cref{sec:related-work}, we summarize the state of the art in object detection and specifically on the detection of solar modules. In \cref{sec:module-detection,sec:crossing-detection}, we introduce our method, which is eventually compare against the state of the art in \cref{sec:experiments}.

\section{Related work}\label{sec:related-work}

The detection of solar modules in an EL image is an object detection task. Traditionally, feature-based methods have been applied to solve the task of object detection. Especially, Haar wavelets have proven to be successful~\cite{papageorgiou1998general}. For an the efficient computation, Viola and Jones~\cite{viola2001rapid} made use of integral images, previously known as summed area tables~\cite{crow1984summed}. Integral images are also an essential part of our method.

The detection of solar modules is related to the detection of checkerboard calibration patterns in the image, since both are planar objects with a regular structure. Recently, integral images have been used with a model-driven approach to robustly and accurately detect checkerboard calibration patterns in presence of blur and noise~\cite{hoffmann2017robust}. We will employ a similar model-driven approach that exploits the regular structure of the cells, but only uses 1-D image statistics. Similar techniques are applied by the document analysis community to detect text lines~\cite{likforman2007text}.

In the last years, convolutional neural networks (CNNs) have achieved superior performance in many computer vision tasks. For example, single-stage detectors like YOLO~\cite{redmon2016you} yield good detection performance with a tolerable computational cost. Multi-stage object detectors, such as R-CNN~\cite{girshick2014rich}, achieve even better results but come with an increased computational cost. In contrast to CNN-based approaches, the proposed method does not require any training data and is computationally very efficient.
% The performance of CNN-based approaches can be increased by using an architecture that provides a good trade-off between accuracy and computational cost~\cite{iandola2016squeezenet}. However, the computational cost introduced by those architectures is still too high to be applicable in real-time on a low-cost CPU. In addition, they usually require a large number of annotated training examples to achieve good performance, although this is partially overcome by transfer-learning~\cite{bengio2012deep} or meta-learning~\cite{finn2017model}.

There are not many preliminary works on the automated detection of solar modules. \etal{Vetter}~\cite{vetter2016automatized} proposed an object detection pipeline that consists of several stacked filters followed by a Hough transform to detect solar modules in noisy infrared thermography measurements. Recently, \etal{Deitsch}~\cite{deitsch2018segmentation} proposed a processing pipeline for solar modules that jointly detects the modules in an EL image, estimates the configuration (\ie the number of rows and columns of cells), estimates the lens distortion and performs segmentation into rectified cell images. Their approach consists of a preprocessing step, where a multiscale vesselness filter~\cite{frangi1998multiscale} is used to extract ridges (separating lines between cells) and bus bars. Then, parabolic curves are fitted onto the result to obtain a parametric model of the module. Finally, the distortion is estimated and module corners are extracted. Since this is, to the best of our knowledge, the only method that automatically detects solar modules and cell crossing points in EL images, we use this as a reference method to assess the performance of our approach.

\section{Detection of the module}\label{sec:module-detection}

This work is supposed to be used for EL images of solar modules in different constellations. As shown in \cref{fig:resultimgs}, modules might be imaged from different viewpoints. In addition, there might be more than one module visible in the image. In this work, we focus on cases, where one module is fully visible and others might be partially viewed, since this commonly happens, when EL images of modules mounted next to each other are captured in the field. However, this method can be easily adapted to robustly handle different situations. The only assumption we make is that the number of cells in a row and per column is known.

The detection of the module in the image and the localization of crossing points between solar cells is performed in two steps. First, the module is roughly located to obtain an initial guess of a rigid transformation between model and image coordinates. 
We describe the procedure in \cref{subsec:module-detection1} and \cref{subsec:module-detection2}. Then, the resulting transform is used to predict coarse locations of crossing points. These locations are then refined as described in \cref{sec:crossing-detection}.

\subsection{Detection of a single module}\label{subsec:module-detection1}

\begin{figure}[t]
    \centering
    \includegraphics[width=.77\linewidth]{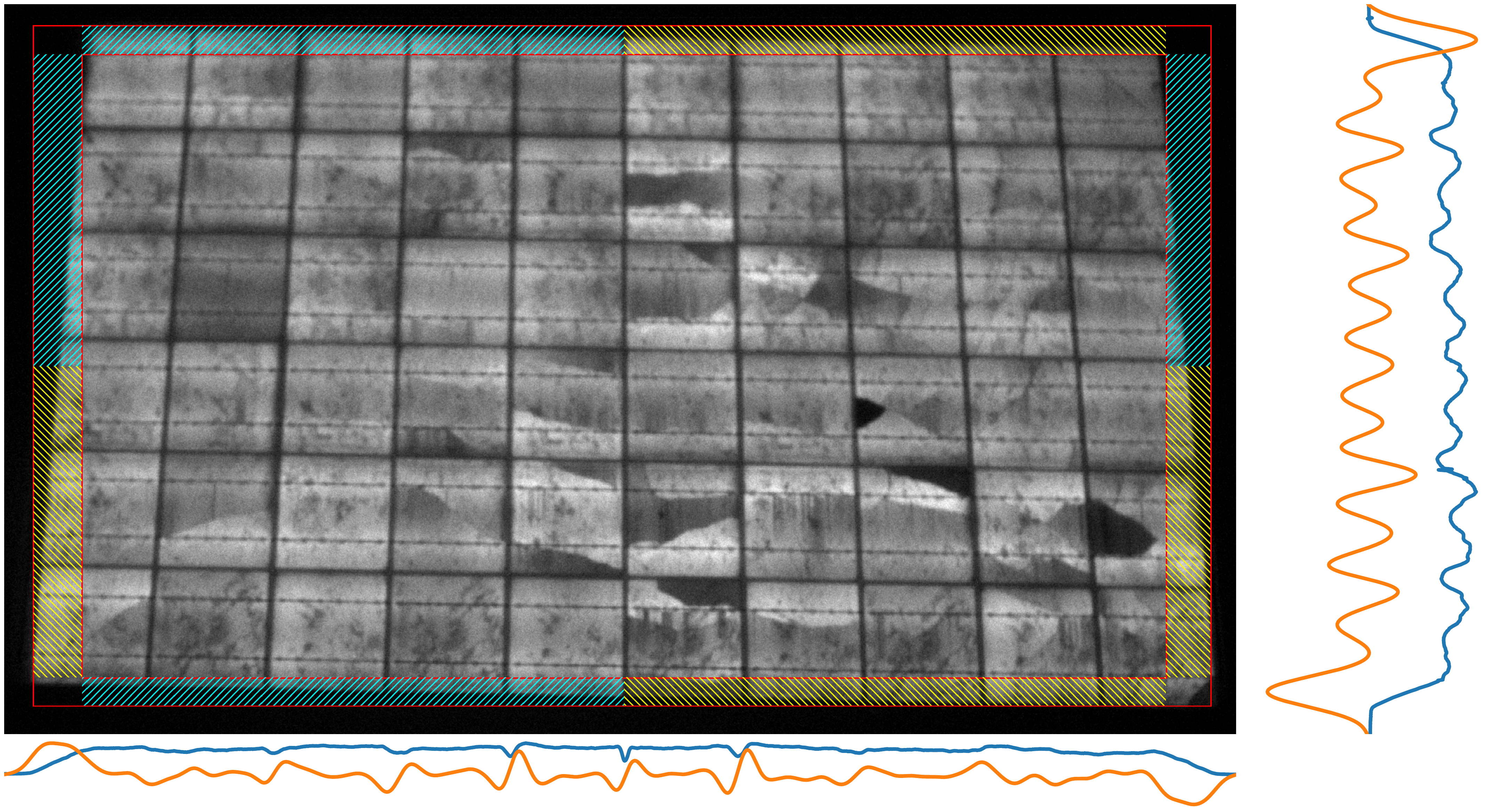}
    \caption{Modules are located by integrating the image in $x$ and $y$ direction (blue lines). From the first derivative of this integration (orange lines), an inner and outer bounding box can be estimated (red boxes). The module corners can be found by considering the pixel sums within the marked subregions.}
    \label{fig:module-detection1}
\end{figure}

We locate the module by considering 1-D images statistics obtained by summing the image in $x$ and $y$ direction. This is related, but not equal to the concept that is known as integral images~\cite{crow1984summed,viola2001rapid}. Let $\img$ denote an EL image of a solar module. Throughout this work, we assume that images are column major, \ie $\img[x,y]$, where $x \in [1,\,w]$ and $y \in [1,\,h]$, denotes a single pixel in column $y$ and row $x$. Then, the integration over rows is given by
\begin{equation}
    \iimgx[y] = \sum_{x=1}^{w} \img[x,y] \;.
\end{equation}
The sum over columns $\iimgy$ is defined similarly. \Cref{fig:module-detection1} visualizes the statistics obtained by this summation (blue lines). Since the module is clearly separated from the background by the mean intensity, the location of the module in the image can be easily obtained from $\iimgx$ and $\iimgy$. However, we are merely interested in the absolute values of the mean intensities than in the change of the latter. Therefore, we consider the gradients $\diimgx$ and $\diimgy$, where $\sigma$ denotes a Gaussian smoothing to suppress high frequencies. Since we are only interested in low frequent changes, we heuristically set $\sigma = 0.01 \cdot \max(w,h)$.

As shown in \cref{fig:module-detection1}, a left edge of a module is characterized by a maximum in $\diimgx$ or $\diimgy$. Similarly, a right edge corresponds to a minimum. In addition, the skewness of the module with respect to the image's $y$ axis corresponds to the width of the minimum and maximum peak in $\diimgx$, whereas the skewness of the module with respect to the $x$ axis corresponds to the peak-widths in $\diimgy$.

Formally, let $x_1$ and $x_2$ denote the location of the maximum and minimum on $\diimgx$, and $y_1$ and $y_2$ denote the location of the maximum and minimum on $\diimgy$, respectively. Further, let $x_{1-}$ and $x_{1+}$ denote the pair of points where the peak corresponding to $x_1$ vanishes. We define two bounding boxes for the module (see \cref{fig:module-detection1}) as follows: The outer bounding box is given by
\begin{equation}
    B_1 = \left[\bc_{1,1},\, \bc_{1,2},\, \bc_{1,3},\, \bc_{1,4}\right]
    = \left[
        (x_{1-},y_{2+}),\,
        (x_{2+},y_{2+}),\,
        (x_{2+},y_{1-}),\,
        (x_{1-},y_{1-})
    \right]\,,
\end{equation}
while the inner bounding box is given by
\begin{equation}
    B_2 = \left[\bc_{2,1},\, \bc_{2,2},\, \bc_{2,3},\, \bc_{2,4}\right]
    = \left[
        (x_{1+},y_{2-}),\,
        (x_{2-},y_{2-}),\,
        (x_{2-},y_{1+}),\,
        (x_{1+},y_{1+})
    \right]\,.
\end{equation}

With these bounding boxes, we obtain a first estimate of the module position. However, it is unclear if $\bc_{1,1}$ or $\bc_{2,1}$ corresponds to the left upper corner of the module. The same holds for $\bc_{1,2}$ versus $\bc_{2,2}$ and so on. This information is lost by the summation over the image. However, we can easily determine the exact pose of the module. To this end, we consider the sum over the sub-regions between the bounding boxes, \cf \cref{fig:module-detection1}. This way, we can identify the four corners $\{\bc_1,\, \ldots,\, \bc_4\}$ of the module and obtain a rough estimate of the module position and pose. To simplify the detection of crossing points, we assume that the longer side of a non-square module always corresponds to the edges $(\bc_1,\,\bc_2)$ and $(\bc_3,\,\bc_4)$.

\subsection{Detection of multiple modules}\label{subsec:module-detection2}

\begin{figure}[tp]
    \centering
    \includegraphics[width=.525\linewidth]{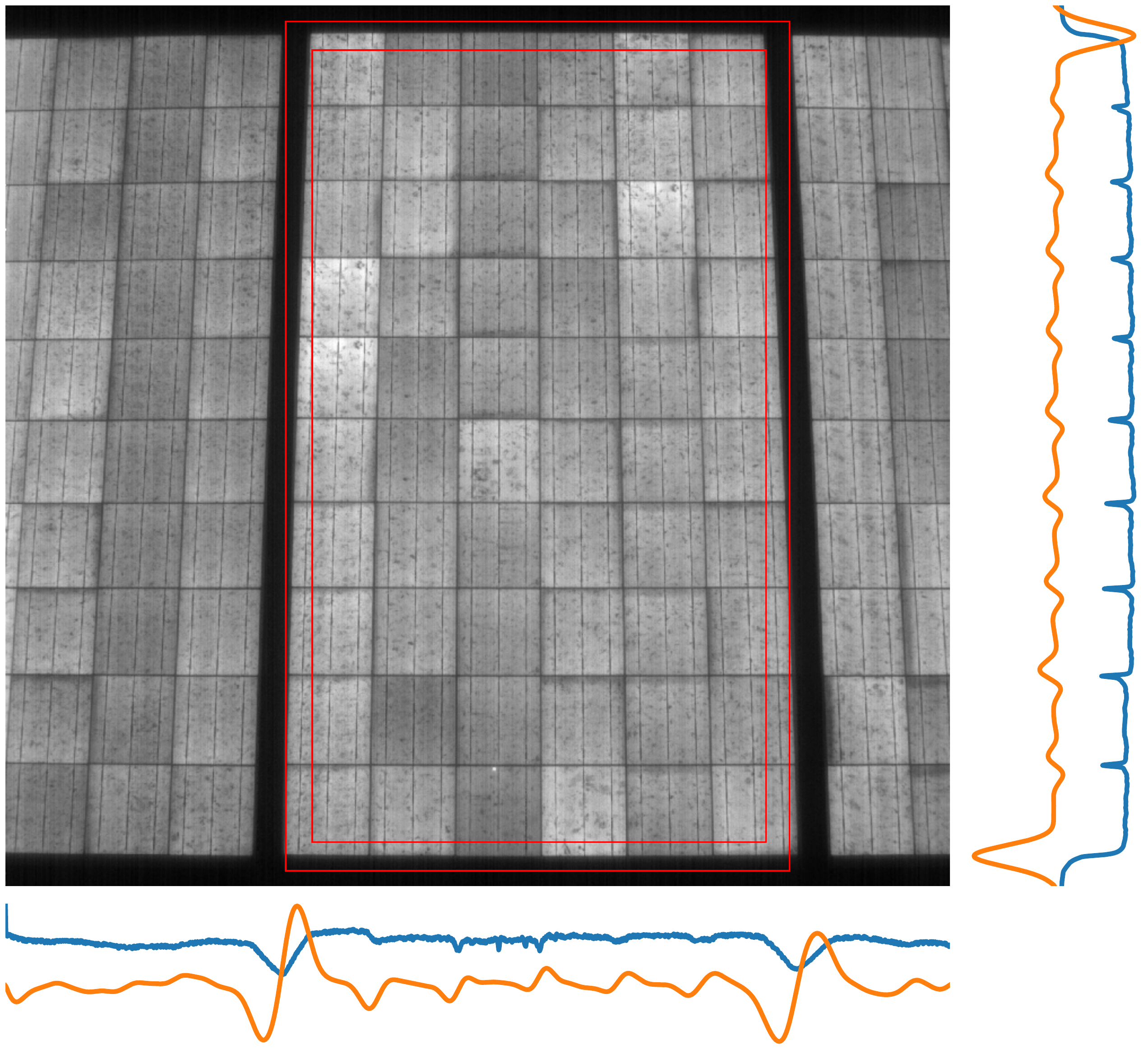}
    \caption{Module detection with multiple modules.}
    \label{fig:module-detection2}
\end{figure}

In many on-site applications, multiple modules will be visible in an EL image (see \cref{fig:module-detection2}). In these cases, the detection of a single maximum and minimum along each axis will not suffice. To account for this, we need to define, when a point in $\diimgk$, $k \in \{x,\,y\}$, will be considered a maximum/minimum. We compute the standard deviation $\sigma_k$ of $\diimgk$ and consider every point a maximum, where $2\sigma_k < \diimgk$ and every point a minimum, where $-2\sigma_k > \diimgk$. Then, we apply non maximum/minimum suppression to obtain a single detection per maximum and minimum. As a result, we obtain a sequence of extrema per axis. Ideally, every minimum is directly followed by a maximum. However, due to false positives this is not always the case.

In this work, we focus on the case, where only one module is fully visible, whereas the others are partially occluded. Since we know that a module in the image corresponds to a maximum followed by a minimum, we can easily identify false positives. We group all maxima and minima that occur sequentially and only keep the one that corresponds to the largest or smallest value in $\diimgk$. Still, we might have multiple pairs of maxima followed by a minimum. We choose the one where the distance between minimum and maximum is maximal.

This is a very simple strategy that does not allow to detect more than one module. However, an extension to multiple modules is straightforward.

\section{Detection of cell crossing points}\label{sec:crossing-detection}

For the detection of cell crossing points, we assert that the module consists of $N$ columns of cells and $M$ rows, where a typical module configuration is $N = 10$ and $M = 6$. However, our approach is not limited to that configuration. Without loss of generality, we  assume that $N \geq M$. With this information, we can define a simple model of the module. It consists of the corners and cell crossings on a regular grid, where the cell size is $1$. By definition, the origin of the model coordinate system resides in the upper left corner with the y axis pointing downwards. 
Hence, every point in the model is given by
\begin{equation}
    \mc_{i,j} = (i-1,\, j-1) \quad i \leq N,\, j \leq M \;.
\end{equation}

From the module detection step, we roughly know the four corners $\{\bc_1,\, \ldots,\, \bc_4\}$ of the module that correspond to model points $\{\mc_{1,1},\, \mc_{N,1},\, \mc_{N,M},\, \mc_{1,M}\}$. Here, we assume that the longer side of a non-square module always corresponds to edges $(\bc_1,\,\bc_2)$ and $(\bc_3,\,\bc_4)$, and that $N \geq M$. Note that this does not limit the approach regarding the orientation of the module since, for example, $(\bc_1,\,\bc_2)$ can define a horizontal or vertical line in the image.

We aim to estimate a transform that converts model coordinates $\mc_{i,j}$ into image coordinates $\xc_{i,j}$, which is done by using a homography matrix $\homo_0$ that encodes the relation between model and image plane. With the four correspondences between the module edges in model and image plane, we estimate $\homo_0$ using the direct linear transform (DLT)~\cite{hartley2003multiple}. Using $\homo_0$, we obtain an initial guess to the position of each crossing point by
\begin{equation}\label{eq:x-approx}
    \xch_{i,j} \approx \homo_0 \mch_{i,j} \;,
\end{equation}
where the model point $\mc = (x,\,y)$ in cartesian coordinates is converted to its homogeneous representation by $\mch = (x,\,y,\,1)$.

Now, we aim to refine this initial guess by a local search. To this end, we extract a rectified image patch of the local neighborhood around each initial guess (\cref{subsec:patches}). Using the resulting image patches, we apply the detection of cell crossing points (\cref{subsec:crossing-detection}). Finally, we detect outliers and re-estimate $\homo_0$ to minimize the reprojection error between detected cell crossing points and the corresponding model points (\cref{subsec:outliers}).

\subsection{Extraction of rectified image patches}\label{subsec:patches}

For the local search, we consider only a small region around the initial guess. By means of the homography $\homo_0$, we have some prior knowledge about the position and pose of the module in the image. We take this into account by warping a region that corresponds to the size of approximately one cell. To this end, we create a regular grid of pixel coordinates. The size of the grid depends on the approximate size of a cell in the image, which is obtained by
\begin{equation}\label{eq:patch-size}
    \hat{r}_{i,j} = \lVert \hat{\xc}_{i,j} - \hat{\xc}_{i+1,j+1} \lVert_2 \;,
\end{equation}
where the approximation $\hat{\xc}$ is given by \cref{eq:x-approx} and conversion from homogeneous $\xch = (x_1,\,x_2,\,x_3)^\tp$ to inhomogeneous coordinates is $\hat{\xc} = \big(\frac{\xh_1}{\xh_3}\,,\frac{\xh_2}{\xh_3}\big)^\tp$. Note that the approximation $\hat{r}_{i,j}$ is only valid in the vicinity of $\hat{\xc}_{i,j}$. The warping is then performed by mapping model coordinates into image coordinates using $\homo_0$ followed by sampling the image using bilinear interpolation. As a result, a rectified patch image $\img_{i,j}$ is obtained that is coarsely centered at the true cell crossing point, see \cref{fig:cell-crossings}.

\subsection{Cell crossing points detection}\label{subsec:crossing-detection}

\begin{figure}[t]
\begin{subfigure}[t]{.33\textwidth}
  \centering
  \includegraphics[width=.8\linewidth]{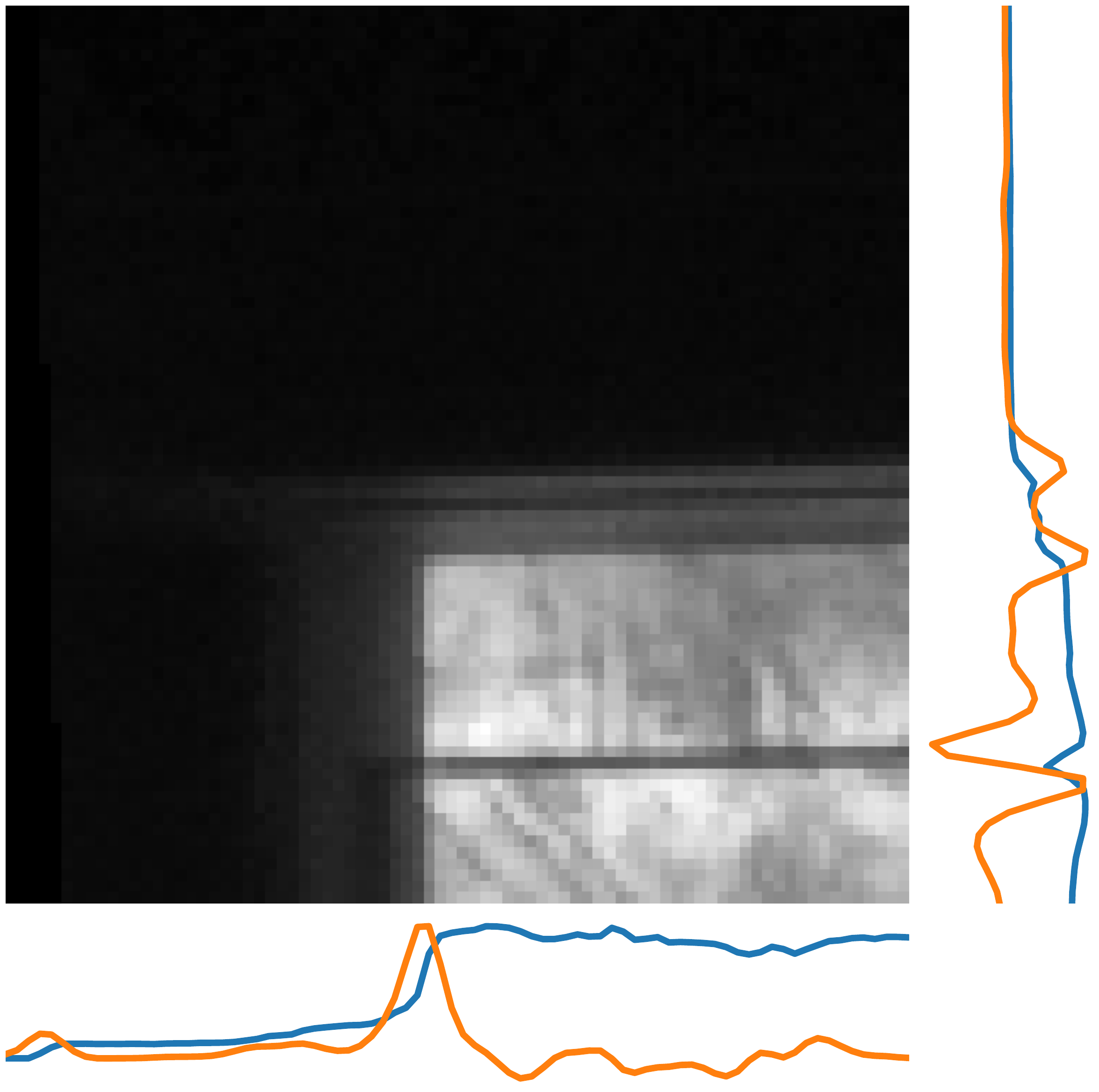}
  \subcaption{Module corner}
  \label{fig:cell-crossings1}
\end{subfigure}%
\begin{subfigure}[t]{.33\textwidth}
  \centering
  \includegraphics[width=.8\linewidth]{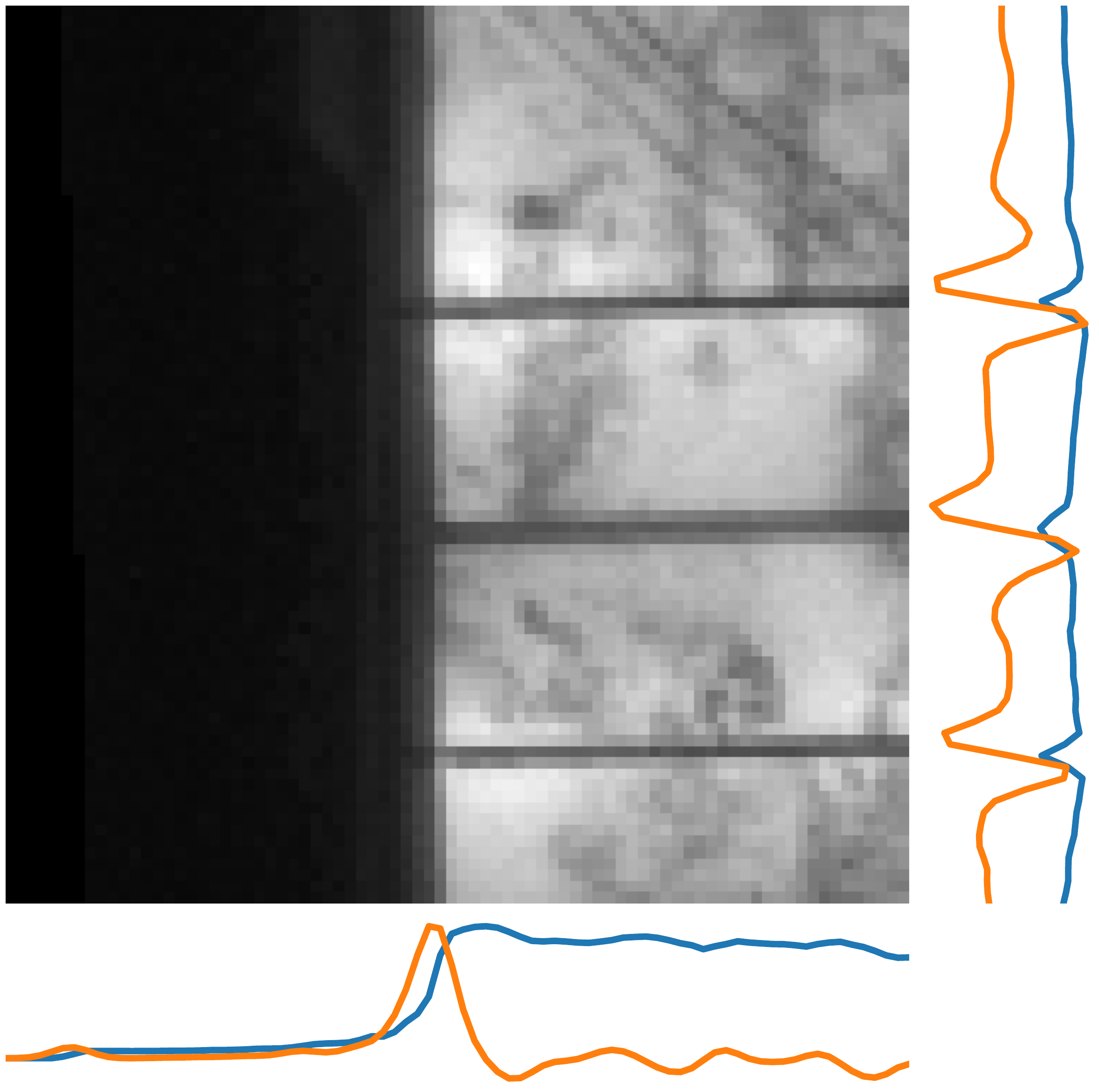}
  \subcaption{Crossing of two cells on an edge of the module}
  \label{fig:cell-crossings2}
\end{subfigure}
\begin{subfigure}[t]{.33\textwidth}
  \centering
  \includegraphics[width=.8\linewidth]{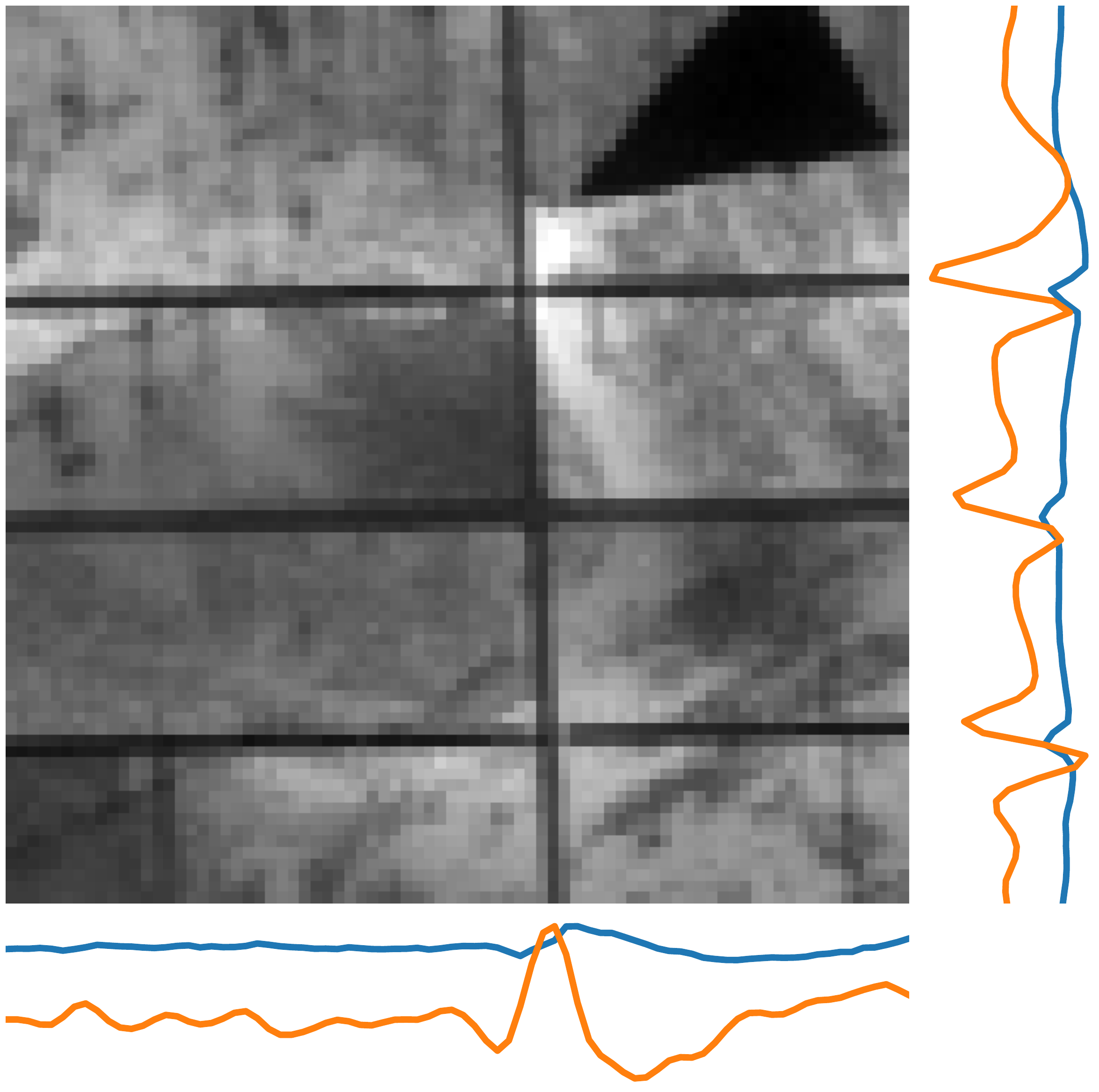}
  \subcaption{Crossing of four cells}
  \label{fig:cell-crossings3}
\end{subfigure}
\caption{Different types of crossings between cells (\cf~\cref{fig:cell-crossings2,fig:cell-crossings3}) as well as the corners of a module (\cf~\cref{fig:cell-crossings1}) lead to different responses in the 1-D statistics. We show the accumulated intensities in blue and the gradient of that in orange.}
\label{fig:cell-crossings}
\end{figure}

The detection step for cell crossing points is very similar to the module detection step but with local image patches. It is carried out for every model point $\mc_{i,j}$ and image patch $\img_{i,j}$ to find an estimate $\xc_{i,j}$ to the (unknown) true image location of $\mc_{i,j}$. To simplify notation, we drop the index throughout this section. We compute 1-D image statistics from $\img$ to obtain $\iimgx$ and $\iimgy$, as well as $\diimgx$ and $\diimgy$, as described in \cref{subsec:module-detection1}. The smoothing factor $\sigma$ is set relative to the image size in the same way as for the module detection.

We find that there are different types of cell crossings that have differing intensity profiles, see \cref{fig:cell-crossings}. Another challenge is that busbars are hard to distinguish from the ridges (separating regions between cells) between cells, see for example \cref{fig:cell-crossings3}. Therefore, we cannot consider a single minimum/maximum. We proceed similar to the approach for the detection of multiple modules, \cf \cref{subsec:module-detection2}). We apply thresholding and non-maximum/non-minimum suppression on $\diimgx$ and $\diimgy$ to obtain a sequence of maxima and minima along each axis. The threshold is set to $1.5\cdot\sigma_k$, where $\sigma_k$ is the standard deviation of $\diimgk$. From the location of $\mc$ in the model grid, we know the type of the target cell crossing. We distinguish between ridges and edges of the module. A cell crossing might consist of both. For example a crossing between two cells on the left border of the module, see \cref{fig:cell-crossings2}, consists of an edge on the $x$ axis and a ridge on the $y$ axis.

\subsubsection{Detection of ridges}

A ridge is characterized by a minimum in $\diimgk$ followed by a maximum. As noted earlier, ridges are hard to distinguish from busbars. Luckily, solar cells are usually built symmetrically. Hence, given that image patches are roughly rectified and that the initial guess to the crossing point is not close to the border of the image patch, it is likely that we observe an even number of busbars. As a consequence, we simply use all minima that are directly followed by a maximum, order them by their position and take the middle. We expect to have an odd number of such sequences (an even number of busbars and the actual ridge we are interested in). In case this heuristic is not applicable, because we found an even number of such sequences, we simply drop this point. The correct position on the respective axis corresponds to the turning point of $\diimgk$.

\subsubsection{Detection of edges}

For edges, we distinguish between left/top edges and bottom/right edges of the module. Left/top edges are characterized by a maximum, whereas bottom/right edges correspond to a minimum in $\diimgk$. In case of multiple extrema, we make a heuristic to choose the correct one. We assume that our initial guess is not far off. Therefore, we choose the maximum or minimum that is closest to the center of the patch.

\subsection{Outlier detection}\label{subsec:outliers}

We chose to apply a fast method to detect the crossing points by considering 1-D image statistics only. As a result, the detected crossing points contain a significant number of outliers. In addition, every detected crossing point exhibits some measurement error. Therefore, we need to identify outliers and find a robust estimate to $\homo$ that minimizes the overall error. Since $\homo$ has $8$ degrees of freedom, only four point correspondences $(\mc_{i,j},\, \hat{\xc}_{i,j})$ are required to obtain a unique solution. On the other hand, a typical module with $10$ rows and $6$ columns has $77$ crossing points. Hence, even if the detection of crossing points failed in a significant number of cases, the number of point correspondences is typically much larger than $4$. Therefore, this problem is well suited to be solved by Random Sample Consensus (RANSAC)~\cite{fischler1981random}. We apply RANSAC to find those point correspondences that give the most consistent model. At every iteration $t$, we randomly sample four point correspondences and estimate $\homo_t$ using the DLT. For the determination of the consensus set, we treat a point as an outlier if the detected point $\xc_{i,j}$ and the estimated point $\homo_t\mch_{i,j}$ differ by more than \SI{5}{\percent} of the cell size.

The error of the model $\homo_t$ is given by the following least-squares formulation
\begin{equation}
    e_t = \frac{1}{NM} \sum_{i,j} \lVert \hat{\xc}_{i,j} - \xc_{i,j} \rVert_2^2 \;,
\end{equation}
where $\hat{\xc}$ is the current estimate by the model $\homo$ in cartesian coordinates. Finally, we estimate $\homo$ using all point correspondences from the consensus set to minimize $e_t$.

%\subsection{Efficient implementation}
%
%\begin{enumerate}
%    \item real vs 1-D integral image
%    \item computation of sums in module detection
%\end{enumerate}

\section{Experimental results}\label{sec:experiments}

%\begin{figure}[t]

%\begin{subfigure}{.33\textwidth}
%  \centering
%  \includegraphics[width=\linewidth,angle=90,trim=40 10 10 70,clip,origin=c]{figures/datasets/deitsch_testset/Result_of_89457_H.jpg}
%  \subcaption{}
%  \label{fig:dataset-examples1}
%\end{subfigure}
%\begin{subfigure}{.33\textwidth}
%  \centering
%  \includegraphics[width=\linewidth]{figures/datasets/multiple/181122_iPV40_Frauenberg_1_5A_20s_B8_0008726521_PM062450.jpg}
%  \subcaption{}
%  \label{fig:dataset-examples2}
%\end{subfigure}%
%\begin{subfigure}{.33\textwidth}
%  \centering
%  \includegraphics[width=.7\linewidth]{figures/datasets/rotated/40.jpg}
%  \subcaption{}
%  \label{fig:dataset-examples3}
%\end{subfigure}

%\caption{Example images from the four datasets: \subref{fig:dataset-examples1} example images from \datab, \subref{fig:dataset-examples2} example from the multi-module dataset \datac, \subref{fig:dataset-examples3} example from the dataset with rotated modules \datad.}
%\label{fig:dataset-examples}
%\end{figure}

We conduct a series of experiments to show that our approach is robust \wrt to the position and pose of the module in the image as well as to various degrees of distortion of the modules. In \cref{subsec:dataset}, we introduce the dataset that we use throughout our experiments. In \cref{subsec:detection-results}, we quantitatively compare the results of our approach with our reference method~\cite{deitsch2018segmentation}. In addition, we show that our method robustly handles cases, where multiple modules are visible in the image or the module is perspectively distorted. Finally, in \cref{subsec:computation-time}, we compare the computation time of our approach to the state of the art.

\subsection{Dataset}\label{subsec:dataset}

\etal{Deitsch}~\cite{deitsch2018segmentation} propose a joint detection and segmentation approach for solar modules. In their evaluation, they use two datasets. They report their computational performance on a dataset that consists of \num{44} modules. We will refer to this dataset as \dataa and use it only for the performance evaluation, to obtain results that are easy to compare. In addition, they use a dataset that consists of \num{8} modules to evaluate their segmentation. We will refer to this data as \datab, see \cref{fig:resultimgs-2}. The data is publicly available, which allows for a direct comparison of the two methods. However, since we do not apply a pixelwise segmentation, we could not use the segmentation masks they also provided. To this end, we manually added polygonal annotations, where each corner of the polygon corresponds to one of the corners of the module.

To assess the performance in different settings, we add two additional datasets. One of them consists of \num{10} images with multiple modules visible. We deem this setting important, since in on-site applications, it is difficult to measure only a single module. We will refer to this as \datac. An example is shown in \cref{fig:resultimgs-4}. The other consists of \num{9} images, where the module has been gradually rotated around the $y$-axis with a step size of \SI{10}{\degree} starting at \SI{0}{\degree}. We will refer to this as \datad, see \cref{fig:resultimgs-1}. We manually added polygonal annotations to \datac and \datad, too.

For the EL imaging procedure of \datac and \datad, two different silicon detector CCD cameras with an optical long pass filter have been used. For the different PV module tilting angles (\datad), a Sensovation "coolSamba HR-320" was used, while for the outdoor PV string measurements a Greateyes "GE BI 2048 2048" was employed (\datac).

\subsection{Detection results}\label{subsec:detection-results}

%\begin{figure}[t]
%    \centering
%    \includegraphics[width=.7\linewidth]{figures/results/joined.pdf}
%    \caption{Detection results on different datasets.}
%    \label{fig:results}
%\end{figure}

\begin{figure}[t]
    \centering
    \pgfplotstableread[col sep = comma]{figures/results/joined.csv}\mydata
    \begin{tikzpicture}
    	\begin{axis}[scale only axis, 
    				height=3.25cm, width=0.75\columnwidth,
    			    xlabel={IoU},
    			    ylabel={recall},
    			    no markers,
    			    legend cell align={left},
    			    legend pos = south east,
    			    legend entries={%
    			        \datab~ours ($8.25$),%
    			        \datab~\cite{deitsch2018segmentation} ($7.5$),%
    			        \datac~ours ($9.8$),%
    			        \datac~\cite{deitsch2018segmentation} ($0.0$),%
    			        \datad~ours ($7.7$),%
    			        \datad~\cite{deitsch2018segmentation} ($2.9$)%
    			    },
    			    every axis plot/.append style={thick},
    			    legend pos = south west,
    			    legend style={fill=white, fill opacity=0.8, draw opacity=1, text opacity=1},
    			%xmax=125, xmin=-5,
    			%restrict x to domain={0:120},
    		]
    		\addplot[red] table[x index = {0}, y index = {1}]{\mydata};
    		\addplot[red, dashed] table[x index = {0}, y index = {2}]{\mydata};
    		\addplot[green!60!black] table[x index = {0}, y index = {3}]{\mydata};
    		\addplot[green!60!black, dashed] table[x index = {0}, y index = {4}]{\mydata};
    		\addplot[blue] table[x index = {0}, y index = {5}]{\mydata};
    		\addplot[blue, dashed] table[x index = {0}, y index = {6}]{\mydata};
    	\end{axis}
    \end{tikzpicture}
    \caption{Detection results on different datasets. We report the AUC in brackets.}
    \label{fig:results}
\end{figure}
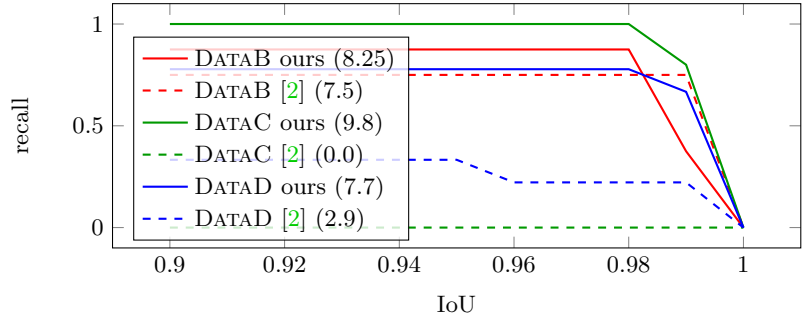

\begin{figure}[t]
    \begin{subfigure}[t]{.23\textwidth}
        \centering
        \includegraphics[height=140pt,trim=40 0 30 0,clip]{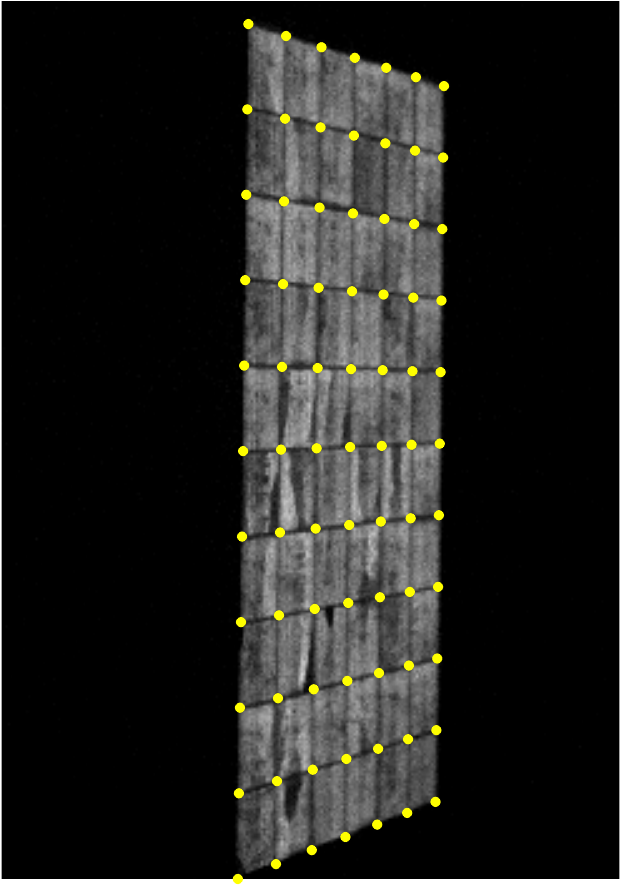}
        \subcaption{}
        \label{fig:resultimgs-1}
    \end{subfigure}
    \begin{subfigure}[t]{.23\textwidth}
        \centering 
        \includegraphics[height=140pt]{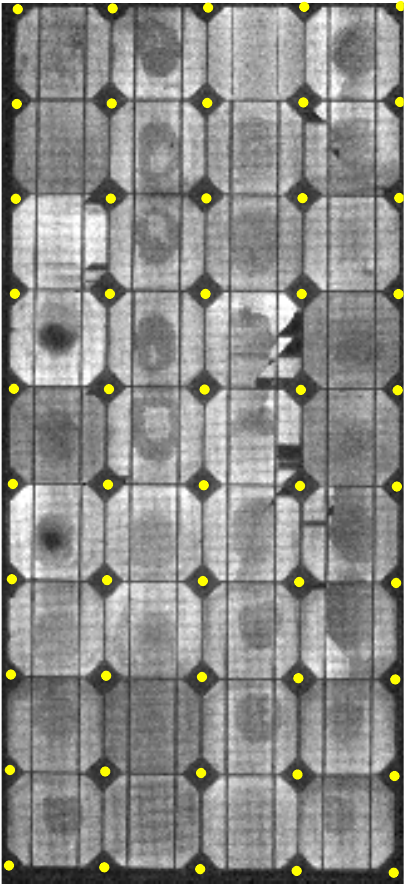}
        \subcaption{}
        \label{fig:resultimgs-2}
    \end{subfigure}
    \begin{subfigure}[t]{.54\textwidth}
        \centering
        \includegraphics[height=140pt,trim=0 30 0 10,clip]{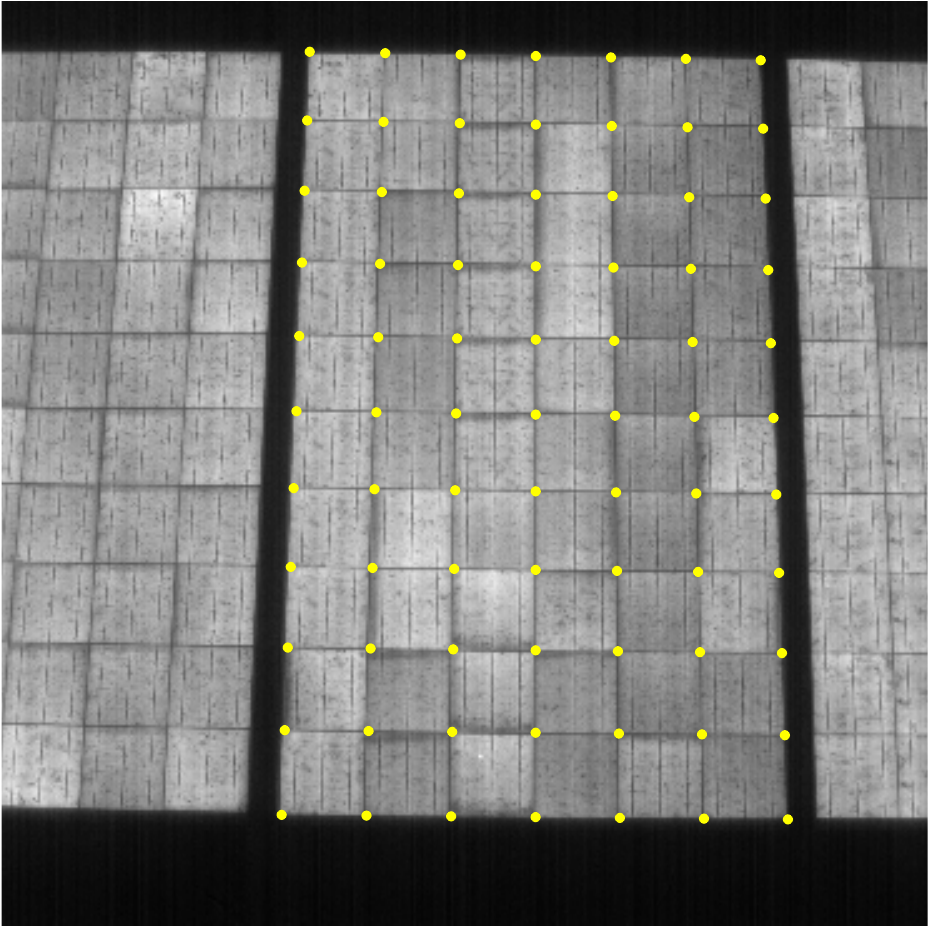}
        \subcaption{}
        \label{fig:resultimgs-4}
    \end{subfigure}

    \caption{Estimated model coordinates on different modules.}
    \label{fig:resultimgs}
\end{figure}
We are interested in the number of modules that are detected correctly and how accurate the detection is. To assess the detection accuracy, we calculate the intersection over union (IoU) between ground truth polygon and detection. Additionally, we report the recall at different IoU-thresholds.

\Cref{fig:results} summarizes the detection results. We see that our method outperforms the reference method on the test dataset provided by \etal{Deitsch}~\cite{deitsch2018segmentation} (\datab) by a small margin. However, the results of the reference method are a little bit more accurate. This can be explained by the fact that they consider lens distortion, while our method only estimates a projective transformation between model and image coordinates. The experiments on \datad assess the robustness of both methods with respect to rotations of the module. We clearly see that our method is considerably robust against rotations, while the reference method requires that the modules are roughly rectified. Finally, we determine the performance of our method, when multiple modules are visible in the image (\datac). The reference method does not support this scenario. It turns out that our method gives very good results when an image shows multiple modules.

In \cref{fig:resultimgs}, we visually show the module crossing points estimated using our method. For the rotated modules (\datad), it turns out that the detection fails for \SI{70}{\degree} and \SI{80}{\degree} rotation. However, for \SI{60}{\degree} and less, we consistently achieve good results (see \cref{fig:resultimgs-2}). Finally, \cref{fig:resultimgs-4} reveals that the method also works on varying types of modules and in presence of severe degradation.

\subsection{Computation time}\label{subsec:computation-time}

We determine the computational performance of our method on a workstation equipped with an Intel Xeon E5-1630 CPU running at \SI{3.7}{\giga\hertz}. The method is implemented in Python3 using NumPy and only uses a single thread. We use the same \num{44} module images that \etal{Deitsch}~\cite{deitsch2018segmentation} have used for their performance evaluation to obtain results that can be compared easily. On average, the \num{44} images are processed in \SI{15}{s}, resulting in approximately \SI{340}{\milli\second} per module. This includes the initialization time of the interpreter and the time for loading the images. The average raw processing time of a single image is about \SI{190}{\milli\second}.

\etal{Deitsch}~\cite{deitsch2018segmentation} report an overall processing time of \SI{6}{\minute} for the \num{44} images using a multi-threaded implementation. Therefore, a single image amounts to \SI{13.5}{\second} on average. Hence, our method is about \num{40} times faster than the reference method. On the other hand, the reference method does not only detect the cell crossing points but also performs segmentation of the active cell area. In addition, they account for lens distortion as well. This partially justifies the performance difference.

\section{Conclusion}

In this work, we have presented a new approach to detect solar modules in EL images. It is based on 1-D image statistics and relates to object detection methods based on integral images. To this end, it can be implemented efficiently and we are confident, that a real-time processing of images is feasible. The experiments show that our method is superior in presence of perspective distortion while performing similarly well than state of the art on non-distorted EL images. Additionally, we show that it is able to deal with scenarios, where multiple modules are present in the image.

In future, the method could be extented to account for complex scenarios, where perspective distortion is strong. In these situations, the stability could be improved by a prior rectification of the module, \eg using the Hough transform to detect the orientation of the module. Since point correspondences between the module and a virtual model of the latter are established, the proposed method could be extended to calibrate the parameters of a camera model, too. This would allow to take lens distortion into account and to extract undistorted cell images.

\section*{Acknowledgements}
We gratefully acknowledge funding of the Federal Ministry for Economic Affairs and Energy (BMWi: Grant No. 0324286, iPV4.0) and  the Erlangen Graduate School in Advanced Optical Technologies (SAOT) by the German Research Foundation (DFG) in the framework of the German excellence initiative.

\bibliographystyle{splncs04}
\bibliography{paper}

\newcommand{\noop}[1]{}
\begin{thebibliography}{10}
\providecommand{\url}[1]{\texttt{#1}}
\providecommand{\urlprefix}{URL }
\providecommand{\doi}[1]{https://doi.org/#1}

\bibitem{crow1984summed}
Crow, F.C.: Summed-area tables for texture mapping. In: ACM SIGGRAPH Computer
  Graphics. pp. 207--212 (1984)

\bibitem{deitsch2018segmentation}
Deitsch, S., Buerhop-Lutz, C., Maier, A., Gallwitz, F., Riess, C.: Segmentation
  of photovoltaic module cells in electroluminescence images. arXiv preprint
  arXiv:1806.06530 [V2]  (2018)

\bibitem{EC2018}
EU energy in figures - statistical pocketbook. European Commission (2018)

\bibitem{fischler1981random}
Fischler, M.A., Bolles, R.C.: Random sample consensus: a paradigm for model
  fitting with applications to image analysis and automated cartography.
  Communications of the ACM  \textbf{24}(6),  381--395 (1981)

\bibitem{frangi1998multiscale}
Frangi, A.F., Niessen, W.J., Vincken, K.L., Viergever, M.A.: Multiscale vessel
  enhancement filtering. In: International Conference on Medical Image
  Computing and Computer-assisted Intervention. pp. 130--137 (1998)

\bibitem{girshick2014rich}
Girshick, R., Donahue, J., Darrell, T., Malik, J.: Rich feature hierarchies for
  accurate object detection and semantic segmentation. In: IEEE Conference on
  Computer Vision and Pattern Recognition. pp. 580--587 (2014)

\bibitem{hartley2003multiple}
Hartley, R., Zisserman, A.: Multiple view geometry in computer vision.
  Cambridge University Press (2003)

\bibitem{hoffmann2017robust}
Hoffmann, M., Ernst, A., Bergen, T., Hettenkofer, S., Garbas, J.U.: A robust
  chessboard detector for geometric camera calibration. In: International
  Conference on Computer Vision Theory and Applications. pp. 34--43 (2017)

\bibitem{likforman2007text}
Likforman-Sulem, L., Zahour, A., Taconet, B.: Text line segmentation of
  historical documents: a survey. International Journal of Document Analysis
  and Recognition (IJDAR)  \textbf{9}(2-4),  123--138 (2007)

\bibitem{papageorgiou1998general}
Papageorgiou, C.P., Oren, M., Poggio, T.: A general framework for object
  detection. In: International Conference on Computer Vision. vol.~6, pp.
  555--562 (1998)

\bibitem{redmon2016you}
Redmon, J., Divvala, S., Girshick, R., Farhadi, A.: You only look once:
  Unified, real-time object detection. In: IEEE Conference on Computer Vision
  and Pattern Recognition. pp. 779--788 (2016)

\bibitem{vetter2016automatized}
Vetter, A., Hepp, J., Brabec, C.J.: Automatized segmentation of photovoltaic
  modules in ir-images with extreme noise. Infrared Physics \& Technology
  \textbf{76},  439--443 (2016)

\bibitem{viola2001rapid}
Viola, P., Jones, M., et~al.: Rapid object detection using a boosted cascade of
  simple features. IEEE Conference on Computer Vision and Pattern Recognition
  \textbf{1},  511--518 (2001)

\bibitem{IEA2018}
Zervos, A. (ed.): Renewables 2018. International Energy Agency (2018)

\end{thebibliography}

\end{document}